\pdfoutput=1
\documentclass[11pt]{article}

\usepackage{ACL2023}

\usepackage{times}
\usepackage{latexsym}
\usepackage{tabularx}
 \usepackage{svg}
\usepackage{amsmath}
\usepackage[T1]{fontenc}
\usepackage[export]{adjustbox}[2011/08/13]
\usepackage[utf8]{inputenc}

\usepackage{microtype}

\usepackage{inconsolata}
\usepackage{booktabs}
\usepackage{multirow}
\usepackage{makecell}
\usepackage{threeparttable}
\usepackage{xcolor,colortbl}
\usepackage{color,soul}
\usepackage{adjustbox}
\usepackage{graphicx}
\usepackage{algorithm}
\usepackage{algpseudocode}
\usepackage{pgf}
\title{Linking in the Long Tail: A Two-Pronged Approach to Event Coreference Resolution}
\title{From Complexity to Clarity: Decomposing Event Coreference Resolution into Two Tractable Problems}
\title{$2 * n$ is better than $n^2$: Decomposing Event Coreference Resolution into Two Tractable Problems}

\author{Shafiuddin Rehan Ahmed$^{1}$\hspace{0.5em} Abhijnan Nath$^{2}$\hspace{0.5em} James H. Martin$^{1}$\hspace{0.5em} Nikhil Krishnaswamy$^{2}$
\vspace{6pt}\\
    $^1$Department of Computer Science, University of Colorado, Boulder, CO, USA \\
    {$ \mathtt{\{shah7567,james.martin\}@colorado.edu}$} \vspace{6pt}\\
    $^2$Department of Computer Science, Colorado State University, Fort Collins, CO, USA\\
    { $ \mathtt{\{abhijnan.nath,nkrishna\}@colostate.edu}$}
}

\begin{document}
\maketitle

% \begin{abstract}
% Event coreference resolution (ECR) is the task of linking mentions of the same event either within or across documents in a pairwise manner. A large majority of mention pairs in any dataset are not coreferent, and many of the pairs that are can be identified using simple techniques such as lemma matching of the event triggers or looking for similar sentences in which the mentions appear. However, there exists a long tail of mention pairs that require deeper reasoning to make linking decisions. Existing methods treat both types of examples as the same when training coreference systems. As a result, models are biased toward surface-level discrimination. To alleviate this bias, we break ECR into two problems. First, we train a model to distinguish similar-looking mentions based on surface-level heuristics. Second, we train a model that learns to find coreferring mentions when the sentence in itself is not enough. Finally, we create the event clusters by merging the decisions from the two models. We show that ...
% \end{abstract}

\begin{abstract}
Event Coreference Resolution (ECR) is the task of linking mentions of the same event either within or across documents. Most mention pairs are not coreferent, yet many that are coreferent can be identified through simple techniques such as lemma matching of the event triggers or the sentences in which they appear. Existing methods for training coreference systems sample from a largely skewed distribution, making it difficult for the algorithm to learn coreference beyond surface matching. Additionally, these methods are intractable because of the quadratic operations needed.
 To address these challenges, we break the problem of ECR into two parts: a) a heuristic to efficiently filter out a large number of non-coreferent pairs, and b) a training approach on a balanced set of coreferent and non-coreferent mention pairs. By following this approach, we show that we get comparable results to the state of the art on two popular ECR datasets while significantly reducing compute requirements. We also analyze the mention pairs that are "hard" to accurately classify as coreferent or non-coreferent\footnote{code repo:  $\mathtt{github.com/ahmeshaf/lemma\_ce\_coref}$}.
\end{abstract}

\newcommand{\Ptp}{\ensuremath{\text{P}^{+}_{\mathtt{easy}}}}
\newcommand{\Pfp}{\ensuremath{\text{P}^{-}_{\mathtt{hard}}}}
\newcommand{\Ptn}{\ensuremath{\text{P}^{-}_{\mathtt{TN}}}}
\newcommand{\Pfn}{\ensuremath{\text{P}^{+}_{\mathtt{FN}}}}

\section{Introduction}
% \vspace*{-2mm}
% TODO: show an easy and a hard example from the ecb dataset
Event coreference resolution (ECR) is the task of finding mentions of the same event within the same document (known as ``within-document coreference resolution,'' or {\it WDCR}) or across text (known as ``cross-document coreference resolution,'' or {\it CDCR}) documents. This task is used for knowledge graph construction, event salience detection and question answering \cite{postma-etal-2018-semeval}.

% (citation need here; a lot of paper say ECR is used for QA without citing. I think there is an opportunity here to fill a gap).

\begin{figure*}
  \centering
  \includegraphics[width=1.0\textwidth]{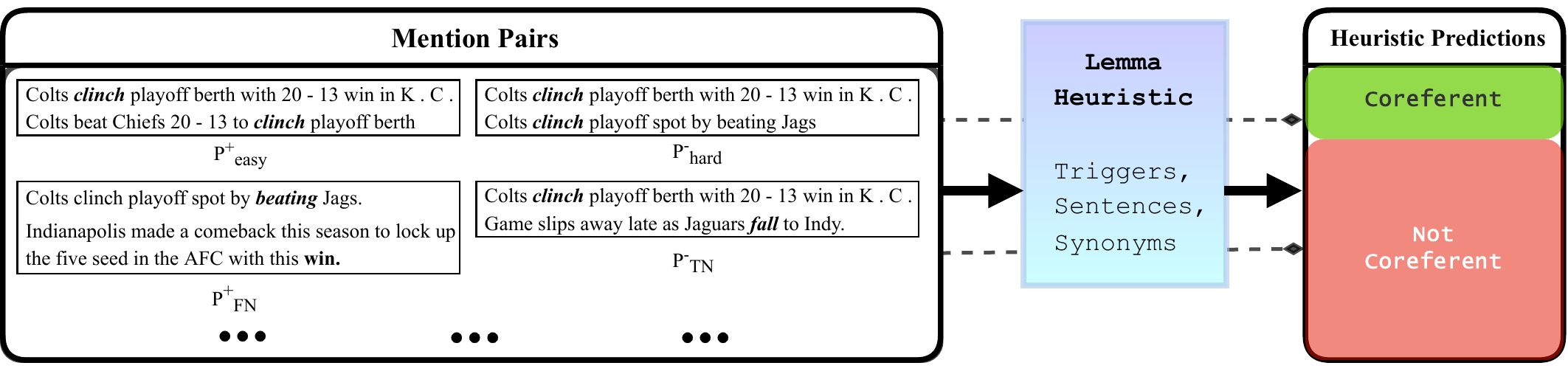}
  \vspace*{-2mm}
\caption{In this approach, we use a lemma-based heuristic to identify coreference, or the relationship between two mentions in a text that refer to the same event. We compare the similarity between the event trigger, which is highlighted in bold and italic, and the lemmas, or base forms, of the sentences. The heuristic classifies the mention pairs "\Ptp" and "\Pfp" as coreferent, and "\Pfn" and "\Ptn" as not coreferent. "\Ptp" and "\Ptn" are correct predictions, meaning they are classified correctly as coreferent and not coreferent. "\Pfp" and "\Pfn" are incorrect predictions, meaning they are misclassified as coreferent and not coreferent.}  \label{fig:Heuristic}
  \vspace*{-2mm}
\end{figure*}

Traditionally, ECR is performed on pairs of event mentions by calculating the similarity between them and subsequently using a clustering algorithm to identify ECR relations through transitivity. The pairwise similarity is estimated using a supervised machine learning method, where an algorithm is trained to distinguish between positive and negative examples based on ground truth. The positive examples are all pairs of coreferent mentions, while the negative examples are all pairs of non-coreferent mentions. To avoid comparing completely unrelated events, the negative pairs are only selected from documents coming from the set of related topics.

% In most cases, references within the same document that refer to the same event do so using the same lemma.

% TODO: concisely come up with the transition.
\newcommand{\Ppos}{\ensuremath{\text{P}^{+}}}
\newcommand{\Pneg}{\ensuremath{\text{P}^{-}}}
Many coreferent pairs are similar on the surface, meaning that the event triggers (the words or phrases referring to the event) have the same lemma and appear in similar sentences. We can use these features in a heuristic to further classify the positive (\Ppos) and negative (\Pneg) pairs into four categories:

\begin{enumerate}
    \vspace*{-2mm}
    \item \Ptp: coreferent/positive mention pairs with high surface similarity.
    \vspace*{-2mm}
    \item \Pfn: coreferent/positive mention pairs with low surface similarity.
    \vspace*{-2mm}
    \item \Pfp: non-coreferent/negative mention pairs with high surface similarity.
    \vspace*{-2mm}
    \item \Ptn: non-coreferent/negative mention pairs with low surface similarity
    \vspace*{-2mm}
\end{enumerate}
% 1) True Positive (\Ptp): positive mention pair with high surface similarity, 2) False Positive (\Pfp): negative mention pair with high surface similarity, 3) False Negative (\Pfn): positive mention pair with low surface similarity, and 4) True Negative (\Ptn): negative mention pair with low surface similarity.

\noindent As shown in Figure \ref{fig:Heuristic}, \Ptp~represents coreferent mention pairs that can be correctly identified by the heuristic, but \Pfp~are non-coreferent pairs that might be difficult for the heuristic to identify. Similarly, \Ptn~(True Negatives) are non-coreferent pairs  that the heuristic can correctly infer, but \Pfn~(False Negatives) require additional reasoning (that {\it Indianapolis} is coreferent with {\it Colts}) to make the coreference judgement.

Most mention pairs are non-coreferent, comprising all pairs corresponding to \Pfp~and \Ptn. However, we observe that that the distribution of the three categories (\Ptp, \Pfp, and \Pfn) is fairly similar across most ECR datasets, with \Ptn~causing the imbalance between positive and negative pairs. Previous methods do not differentiate between these four categories and randomly select the positive and negative pairs to train their coreference systems from this heavily skewed distribution.  This makes it challenging for the coreference algorithm to identify coreferent links among a large number of non-coreferent ones. Furthermore, as ECR is performed on $n^2$ number of mention pairs, where $n$ is the number of mentions in the corpus, these methods can become intractable for a large corpus.

% Out of the four categories of mention pairs, we observe that the distribution of the three categories (\Ptp, \Pfp, and \Pfn) is relatively similar, while the fourth category (\Ptn) is significantly larger in size. This leads to a significant imbalance between the number of pairs that are coreferent and those that are not. Previous methods select positive and negative examples from this heavily skewed distribution, which makes it very difficult for the coreference algorithm to learn how to identify coreferent links among a large number of non-coreferent ones.

To improve the efficiency of the ECR process while achieving near sate of the art (SOTA) results, we divide the problem into two manageable subtasks: a) a heuristic to efficiently and accurately filter out a large number of \Ptn~as a way of balancing the skewed distribution, and b) an ECR system trained on the balanced set of coreferent and non-coreferent mention pairs (\Ptp~and \Pfp). This approach also eases the analysis of some of the mention pairs that are difficult to classify with an ECR system, which we present in this paper.

% In the first method, we only select positive and negative samples from pairs of mentions that are very similar in terms of their trigger and sentence structure, specifically the \Ptp~and \Pfp~pairs. This helps the classifier learn to identify non-coreferent pairs that are mistakenly classified as coreferent by a simple heuristic. In the second approach, we only choose samples that are significantly different in context (\Pfn~and \Ptn). This helps the classifier learn to link the long tail of coreferent pairs (\Pfn) that are disregarded by the heuristic. Finally, we take the predictions from the two classifiers and perform clustering to produce the final event clusters.

% We observe ...

\newcommand{\ecb}{ECB+}
\newcommand{\gvc}{GVC}
\vspace*{-2mm}
\section{Related Work}
\vspace*{-2mm}
% TODO: summarize the methods in coreference.bib, {easy_first.bib}
\newcommand{\eA}{\ensuremath{\text{E}_{\text{A}}}}
\newcommand{\eB}{\ensuremath{\text{E}_{\text{B}}}}
\newcommand{\eCLS}{\ensuremath{\text{E}_{\text{CLS}}}}
\newcommand{\CE}{\ensuremath{\texttt{CE}}}

\paragraph{Pre-Transformer Methods}
Pre-Transformer language model-related works in event coreference such as \citet{kenyon2018resolving} trained neural models with customized objective (loss) functions to generate richer representations of mention-pairs using ``static'' embeddings such as contextual Word2Vec \cite{mikolov2013efficient} as well as document-level features such as TF-IDF and heuristically-motivated features like mention-recency, word overlap, and lemma overlap, etc. As such, they improved upon the baselines established by \citet{cybulska-vossen-2015-translating} on the \ecb~corpus. Similarly, works such as \citet{barhom-etal-2019-revisiting} suggest both disjoint and joint-clustering of events mentions with their related entity clusters by using a predicate-argument structure. In this, their disjoint model surpassed \citet{kenyon2018resolving} by 9.5 F1 points using the CoNLL scorer \cite{pradhan-EtAl:2014:P14-2} whereas their joint model improved upon the disjoint model by 1.2 points for entities and 1 point for events.

\vspace*{-2mm}
\paragraph{Transformer-based Cross-encoding}
Most recent works \cite{meged-etal-2020-paraphrasing,zeng-etal-2020-event,
cattan-etal-2021-cross-document, allaway-etal-2021-sequential, caciularu-etal-2021-cdlm-cross, held-etal-2021-focus,yu-etal-2022-pairwise} in CDCR   have shown  success in using pairwise mention representation learning models, a method popularly known as cross-encoding. These methods use distributed and contextually-enriched ``non-static'' vector representations of mentions from large, Transformer-based language models like various BERT-variants to calculate supervised pairwise scores for those event mentions. At inference, such works use variations of incremental or agglomerative clustering techniques to form predicted coreference links and evaluate their chains on gold coreference standards. The methods vary with the context they use for cross-encoding. \citet{cattan-etal-2021-cross-document} use only sentence-level context, \citet{held-etal-2021-focus} use context from sentences surrounding the mentions, and \citet{caciularu-etal-2021-cdlm-cross} use context from entire documents.

In our research, we have focused on the CDLM model from \citet{caciularu-etal-2021-cdlm-cross} and their methodology, which uses a combination of enhanced pretraining using the global attention mechanism inspired by \citet{beltagy2020longformer} as well as finetuning on a task-specific dataset using pretrained special tokens to generate more semantically-enhanced embeddings for mentions. \citet{beltagy2020longformer} and \citet{caciularu-etal-2021-cdlm-cross} cleverly use the global attention mechanism to linearly scale the oft-quadratic complexity of pairwise scoring of mentions in coreference resolution while also accommodating longer documents (up to 4,096 tokens).  Previous works such as \citet{baldwin-1997-cogniac}, \citet{stoyanov2012easy}, \citet{lee-etal-2012-joint}, and \citet{lee2013deterministic} also reduce computation time by strategically using deterministic, rule-based systems along with neural architectures.

%Cattan uses sentence-level context
%Held uses context from surrounding sentences (get %details)
%Caciularu uses entire doc
%TODO: Sentence \cite{cattan-etal-2021-cross} vs Surrounding sentece \cite{held-etal-2021-focus} vs Entire doc \cite{caciularu-etal-2021-cdlm-cross}.

%TODO: Training approach\cite{kenyon2018resolving}

%TODO: heuristic + easy first %\cite{stoyanov2012easy}, %\cite{lee2013deterministic}
%\cite{lee-etal-2012-joint} These handle easy examples first with heuristics

Recently, pruning \Ptn~for ECR has been shown to be effective by \citet{held-etal-2021-focus}. They create individual representations for mentions and use them in a bi-encoder method to retrieve potential coreferent candidates, which are later refined using a cross-encoder trained on hard negative examples. In contrast, our approach utilizes a computationally efficient pruning heuristic and trains the cross-encoder on a smaller dataset. We also conduct an error analysis on all hard examples that are misclassified by the cross-encoder, which is made feasible by the heuristic.

\vspace*{-0.5mm}
\section{Datasets}

\newcommand{\ace}{ACE 2005}
\vspace*{-0.5mm}
We experiment with two popular ECR datasets distinguished by the effectiveness of a lemma heuristic on the dataset.
\subsection{Event Coreference Bank Plus (\ecb)}
\vspace*{-0.5mm}
The \ecb~corpus \cite{cybulska-vossen-2014-using} is a popular English corpus used to train and evaluate systems for event coreference resolution. It extends the Event Coref Bank corpus (ECB; \citet{bejan-harabagiu-2010-unsupervised}), with annotations from around 500 additional documents. The corpus includes annotations of text spans that represent events, as well as information about how those events are related through coreference. We divide the documents from topics 1 to 35 into the training and validation sets\footnote{Validation set includes documents from the topics 2, 5, 12, 18, 21, 34, and 35}, and those from 36 to 45 into the test set, following the approach of \citet{cybulska-vossen-2015-translating}.

% \vspace*{-1mm}
\subsection{Gun Violence Corpus (\gvc)}
% \vspace*{-1mm}
% TODO: change this GPT generated descriptions:

% The Gun Violence Corpus \cite{vossen2018don} is a collection of text data related to incidents of gun violence. It may include news articles, social media posts, government reports, and other sources of information about gun violence. The corpus is used for research and analysis, with the goal of understanding and addressing the issue of gun violence in society. The data in the corpus may be annotated and structured in a specific way to facilitate research tasks such as event extraction and coreference resolution. The corpus may be used by researchers in fields such as sociology, criminology, and public health to study the causes and consequences of gun violence and develop interventions to prevent it. Data split by \citet{bugert-etal-2021-generalizing}.

The Gun Violence Corpus \cite{vossen2018don}
 is a recent English corpus exclusively focusing on event coreference resolution. It is intended to be a more challenging dataset than \ecb~which has a very strong lemma baseline \cite{cybulska-vossen-2014-using}.
 \def\arraystretch{1.2}%
\begin{table}[t]
\centering
\small
    \begin{tabular}{|c|c|c|c|c|c|c|}  \cline{2-7}
     \multicolumn{1}{c|}{~} & \multicolumn{3}{c|}{\ecb} & \multicolumn{3}{c|}{\gvc}  \\ \cline{2-7}
    \multicolumn{1}{c|}{~} & Train & Dev & Test & Train & Dev & Test \\ \hline
	T/ST   & 25 & 8 & 10/20 & 1/170 & 1/37 & 1/34 \\ \hline
	D & 594 & 196 & 206 & 358 & 78 & 74  \\ \hline

	M  & 3808 &  1245 & 1780 & 5313 & 977  & 1008\\ \hline

	C & 1464 & 409 & 805 & 991 & 228 &194 \\ \hline

    S & 1053 & 280 & 623 & 252 & 70 & 43 \\ \hline
% 		\makecell{Singleton \\ Event Instances} & 68 & 107 & 89 \\ \hline

	\end{tabular}
  \vspace*{-2mm}
  \caption[\ecb~Corpus Statistics]{
	\ecb~ and GVC Corpus statistics for event mentions. T/ST = topics/sub-topics, D = documents, M = event mentions, C = clusters, S = singletons.}
\label{tab:ecb_gvc}
  \vspace*{-2mm}
\end{table}
 It is a collection of texts surrounding a single topic (gun violence) and various sub-topics. Since it does not have coreference links across sub-topics, we only consider mention pairs within the sub-topics. We use the data split by \citet{bugert-etal-2021-generalizing}. Table \ref{tab:ecb_gvc} contains the statistics for \ecb~and \gvc~corpora.  % \subsection {Dataset Distributions}

\section{System Overview}
\vspace*{-2mm}
There are two major components in our system: the heuristic and the discriminator (cross-encoder) trained on the output of the heuristic.

\newcommand{\LH}{\ensuremath{\texttt{LH}}}
\newcommand{\LHOracle}{\ensuremath{\texttt{LH}_{\texttt{Ora}}}}
\newcommand{\DLH}{\ensuremath{\texttt{LH}_\texttt{D}}}

\vspace*{-1mm}
\subsection{Lemma Heuristics (\LH, \LHOracle)}
% \vspace*{-1mm}
% One unique aspect of ECR is that it has a very high baseline simply by comparing the lemmas of the mention triggers and sentences. We incorporate this feature into our system as the initial step of coreference resolution. We use spaCy\footnote{https://spacy.io/ model \texttt{en\_core\_web\_md} v3.4} to extract the lemmas, a common choice for this task. In addition to matching the lemma of the triggers, we create and use a set of lemma pairs that appear in coreferent mention pairs of the training set.

\newcommand{\tsubA}{\ensuremath{t_{A}}}
\newcommand{\tsubB}{\ensuremath{t_{B}}}
\newcommand{\lsubA}{\ensuremath{l_{A}}}
\newcommand{\lsubB}{\ensuremath{l_{B}}}

A key feature of ECR is its high baseline achieved by comparing the lemmas of mention triggers and sentences. To leverage this feature, we incorporate it as the first step in our coreference resolution system. We utilize spaCy\footnote{https://spacy.io/ model \texttt{en\_core\_web\_md} v3.4} to extract the lemmas, a widely-used tool for this task. In addition to matching lemmas of triggers, we also create and utilize a set of synonymous\footnote{The words need not be synonyms in strict definitions, but rather appear in coreference chains.} lemma pairs that commonly appear in coreferent mention pairs in our training set. This approach allows us to identify coreferent mention pairs that have different triggers and improve the overall recall. The heuristic, \LH, only utilizes the synonymous lemma pairs from the training set. We also evaluate the performance of \LHOracle, which uses synonymous lemma pairs from the entire dataset which means it uses the coreference information of the development and test sets to create synonymous lemma pairs.
\newcommand{\SynP}{\ensuremath{\text{Syn}_{\text{P}}}}

For a mention pair (A, B), with triggers (\tsubA, \tsubB), head lemmas (\lsubA, \lsubB) and for a given synonymous lemma pair set (\SynP), we consider only lemma pairs that pass any of the following rules:
\begin{itemize}
    \vspace*{-2mm}
    \item $ (\lsubA, \lsubB) \in \SynP$
    %\vspace*{-2mm}
    \item $ \lsubA == \lsubB$
    \vspace*{-2mm}
    \item $ \tsubB~contains~\lsubA$
    \vspace*{-2mm}
    \item $ \tsubA~contains~\lsubB$
\end{itemize}
% Additionally, we match the trigger's headword by checking if they are synonyms in WordNet \cite{wordnet}.

For mentions that have matching trigger lemmas/triggers or are synonymous, we proceed by comparing the context of the mentions. In this work, we only compare the mention's sentence to check for similarities between two mentions. To further refine our comparison, we remove stop words and convert the tokens in the text to their base form. Then, we determine the overlap between the two mentions and predict that the pair is coreferent if the overlap exceeds a certain threshold. We tune the threshold using the development sets.

% \vspace*{-2mm}
\begin{figure}[t]
  % \centering
  \includegraphics[width=0.50\textwidth,center]{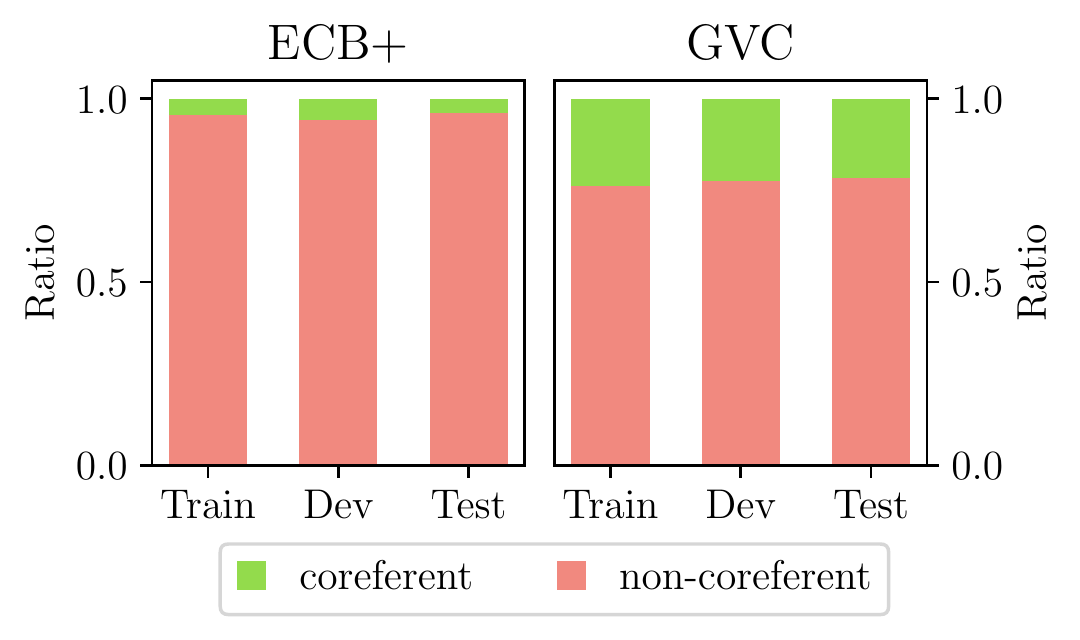}
  \vspace*{-5mm}
\caption{Coreferent vs. non-coreferent mention pairs ratio across datasets.}  \label{fig:mps}
  % \vspace*{-2mm}
\end{figure}
\subsubsection{Filtering out \Ptn}
% \vspace*{-1mm}

Cross-document coreference systems often struggle with a skewed distribution of mention pairs, as seen in Figure \ref{fig:mps}. In any dataset, only 5-10\% of the pairs are corefering, while the remaining 90\% are non-coreferent. To address this, we use the heuristic to balance the distribution by selectively removing non-coreferent pairs (\Ptn), while minimizing the loss of coreferent pairs (\Pfn). We do this by only considering the mention pairs that the heuristic predicts as coreferent, and discarding the non-coreferent ones.

\subsubsection{\Pfp, \Ptp, and \Pfn~Analysis}
% \vspace*{-1mm}

\textbf{\Ptp~and \Pfp}: As defined earlier, \Ptp~are the mention pairs that the heuristic correctly predicts as coreferent when compared to the ground-truth, and \Pfp~are the heuristic's predictions of coreference that are incorrect when compared to the ground-truth. In \S \ref{sec:ptp-pfp}, we go through how we fix heuristic's \Pfp~predictions while minimizing the errors introduced in terms of \Ptp.

\begin{figure}[t]
  \centering
  \includegraphics[width=0.3\textwidth]{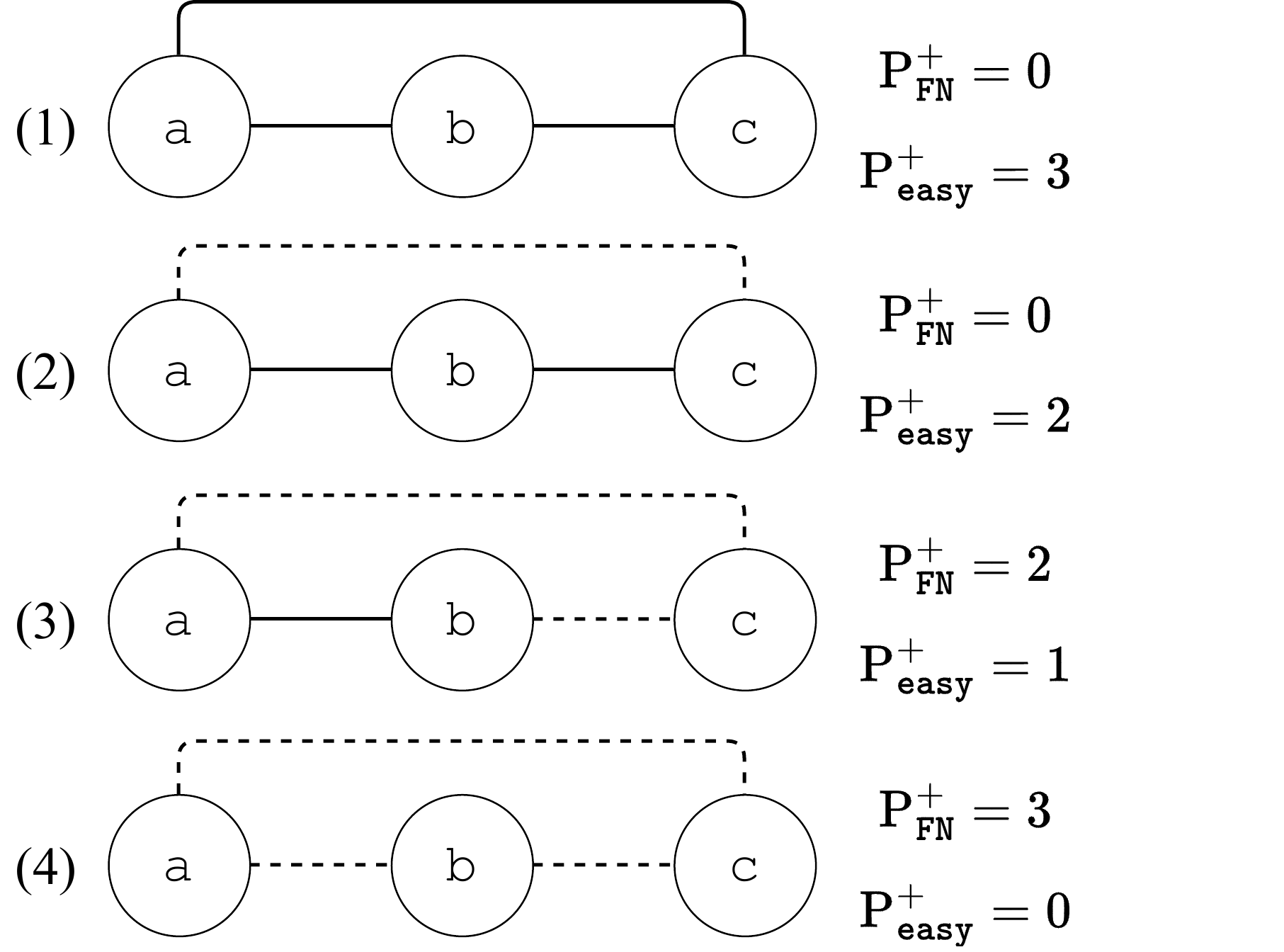}
% \vspace*{-2mm}
\caption{Counting size of mention pairs (\Pfn~and \Ptp) in a true cluster \{a, b, c\} using heuristic's coreferent predictions (solid line) and non-coreferent predictions (dotted line). We count \Pfn~after performing transitive closure, resulting in a size of 0 (instead of 1) in (2).}
\label{fig:fntp}
% \vspace*{-2mm}
\end{figure}

\noindent \textbf{\Pfn}: We define a pair as a  \Pfn~only if it cannot be linked to the true cluster through subsequent steps. As shown in Figure \ref{fig:fntp}, if a true cluster is \{a, b, c\} and the heuristic discards one pair (a, c), it will not be considered as a \Pfn~because the coreference can be inferred through transitivity. However, if it discards two pairs \{(a,c), (b,c)\}, they will both be considered as \Pfn.  We hypothesize that an ideal heuristic is one that maintains a balance between \Ptp~and \Pfp~while minimizing \Pfn, and therefore, we tune the heuristic's threshold accordingly using the development sets of the corpora.
\begin{figure}[b!]
  % \centering
  \includegraphics[width=0.45\textwidth,center]{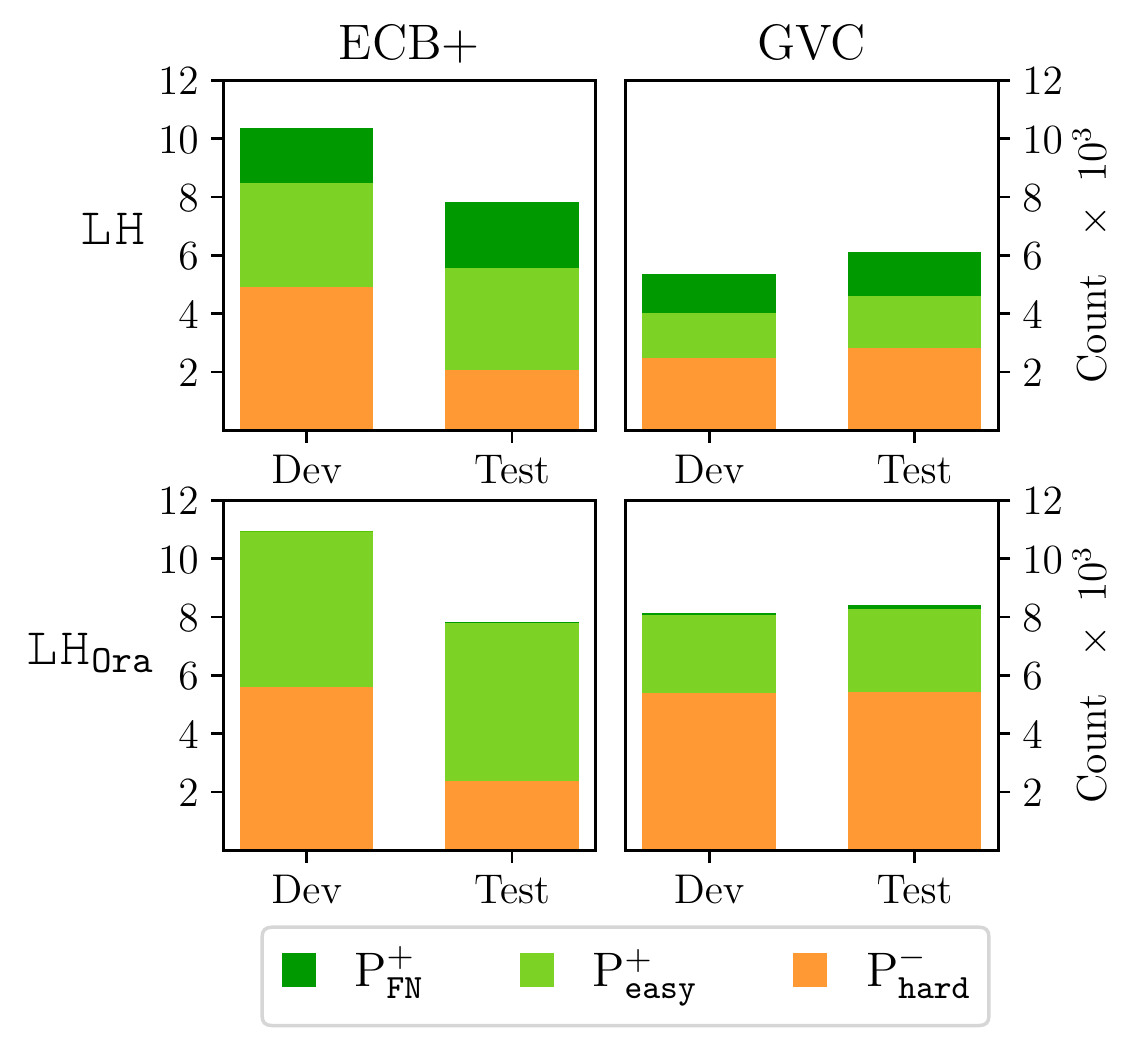}
  % \vspace*{-6mm}
\caption{\LH~and \LHOracle~Distributions of \Pfp, \Ptp, and \Pfn~for \ecb~and \gvc~corpora. \LHOracle~ensures no (or negligible) loss in \Pfn.}  \label{fig:tpfpfn}
  \vspace*{-2mm}
\end{figure}

We evaluate the heuristics \LH~and \LHOracle~by plotting the distributions \Ptp, \Pfp, and \Pfn~generated by each for
 the two corpora. From Figure \ref{fig:tpfpfn}, We observe similar distributions for the test and development sets with the chosen threshold value from the development set. We also observe that \LH~ causes a significant number of \Pfn, while \LHOracle~has a minimal number of \Pfn. Minimizing the count of \Pfn~is important as it directly affects the system's recall. The distributions of \Ptp~and \Pfp~remain balanced across all datasets except when \LHOracle~is used in \gvc~where there are double the number of \Pfp~to \Ptp. \Pfp~should be minimized as it can affect the system's overall precision.
\begin{figure}[t]
  \centering
  \includegraphics[width=0.48\textwidth]{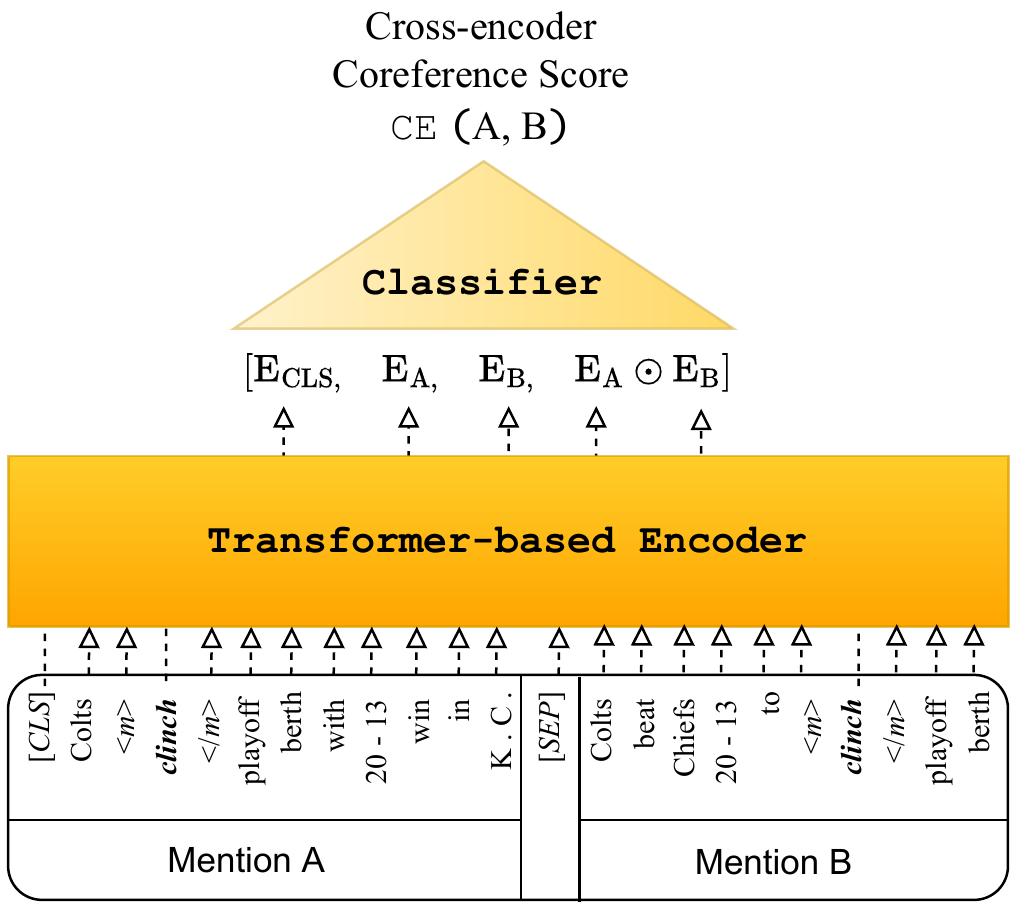}
\vspace*{-2mm}
\caption{The cross-encoding technique to generate the coreference score between the mention pair (A, B). This involves adding special tokens, {\it <m>} and {\it </m>}, around the event triggers, and then combining and processing the two mentions through a transformer-based language model. Certain outputs of the transformer (\eCLS, \eA, \eB) are then concatenated and fed into a classifier, which produces a score between 0 and 1 indicating the degree of coreference between the two mentions.}
\label{fig:crossEncoder}
\vspace*{-2mm}
\end{figure}
%\subsection{\Ptp~and \Pfp~Discriminator}
%\label{sec:ptp-pfp}
% \vspace*{-1mm}
\subsection{Cross-Encoder}
% \vspace*{-1mm}

A common technique to perform ECR is to use Transformer-based cross-encoding (\CE) on the mention pair (A, B). This process, depicted in Figure \ref{fig:crossEncoder}, begins by surrounding the trigger with special tokens (<\textit{m}> and </\textit{m}>). The mentions are then combined into a single input for the transformer (e.g., RoBERTa). The pooled output of the transformer (\eCLS) and the output corresponding to the tokens of the event triggers (\eA~and \eB) are extracted.\footnote{\eA~and \eB~represent the sum of the output embedding of each token for event triggers with multiple tokens.} \eCLS, \eA, \eB, and the element-wise product of the mention embeddings (\eA $\odot$\eB) are all concatenated to create a unified representation of the mention pair. This representation is used, with a classifier, to learn the coreference score, $\CE\left(\text{A, B}\right)$, between the pair after finetuning the transformer.

% \subsection{Easy/Hard Classification}
% As explained earlier, among the mention pairs, there are x \% of those where a simple rule-based heuristic is sufficient to make an accurate coreference decision. We classify this set of mention pairs as Easy. There are also a lot of pairs that need non-trivial reasoning (like entity coreference, synonym, metaphors) to make the coreference decision and we call these examples Hard. We train a model to automatically whether a mention pair is Easy or Hard.

% \subsection{Sampling TP/FP/TN/FN Mention Pairs}
% \subsubsection{Symmetric Scorer}
\newcommand{\tA}{\text{A}}
\newcommand{\tB}{\text{B}}
\newcommand{\tC}{\text{C}}

% \begin{figure}[t]
%   \centering
%   \includegraphics[width=0.445\textwidth, center]{plots/training_samples.pdf}
% \vspace*{-2mm}
% \caption{Training Samples of previous methods vs. ours. The heuristic creates a balanced and significantly smaller training set for \ecb. For \gvc, the heuristic discards half of the negative samples while somewhat balancing the dataset.}  \label{fig:training-pairs}
% \vspace*{-2mm}
% \end{figure}

% \newcommand{\dPos}{\ensuremath{\texttt{D}_{\texttt{POS}}}}
\newcommand{\dPos}{\ensuremath{\texttt{D}}}
\newcommand{\dLong}{\ensuremath{\dPos_{\text{long}}}}
\newcommand{\dSmall}{\ensuremath{\dPos_{\text{small}}}}

% \vspace*{-1mm}
\subsubsection{\Ptp~\& \Pfp~Discriminator (\dPos)}
\label{sec:ptp-pfp}
% \vspace*{-1mm}

The cross-encoder's encoding is non-symmetric, meaning, depending on the order in which the mentions are concatenated, it will give different coreference scores. In reality, the order should not matter for predicting if the two events are the same or not. We propose a symmetric cross-encoding scorer where we take the average of the scores predicted from both combinations of concatenation. So for a mention pair, $p = $ (A, B), the symmetric cross-encoder coreference scorer (\texttt{D}) is given as:

% \vspace*{-2mm}
\begin{equation}
\label{eq:symmetric}
    \texttt{D}(p) = \dfrac{\CE(\tA, \tB) + \CE(\tB, \tA)}{2}
\end{equation}
% \vspace*{-2mm}
We employ a cross-encoder with a symmetric scorer, as outlined in Equation \ref{eq:symmetric}, as the discriminator for \Ptp~and \Pfp. We conduct experiments utilizing two different Transformer models, RoBERTa (\dSmall) and Longformer (\dLong), which vary in their maximum input capacity.

% However, unlike the previous cross encoder architectures that learn to score the mentions in a non-symmetric and one-directional manner, our first-stage discriminator learns to score the mentions while maintaining commutative symmetry. In other words, it uses a linear combination of the commutative binary-loss of the bi-directional mention embeddings, i.e., for a sample pair, AB and BA, it minimizes the combined binary cross entropy losses of AB as well as BA. We train the network with a batch-size of 30 for 15 iterations with a discriminator learning rate of 0.0001 and a Longformer learning rate (fine tuning) of 0.00001.  We use both Longformer and RoBERTa-{\sc base}.

% \vspace*{-2mm}

% \vspace*{-2mm}

\newcommand{\dNeg}{\ensuremath{\texttt{D}_{\texttt{NEG}}}}
\newcommand{\Pall}{\ensuremath{\text{P}_{\text{all}}}}
\newcommand{\Mv}{\ensuremath{M^{v}}}

% \vspace*{-2mm}
\section{Experimental Setup}
% \vspace*{-2mm}
We describe our process of training, prediction, and hyperparameter choice in this section.

% \vspace*{-1mm}
\subsection{Mention Pair Generation}
% \vspace*{-1mm}

We use the gold mentions from the datasets. Following previous methods, we generate all the pairs (\Pall) of mentions (\Mv) from documents coming from the same topic.
We use gold topics in the training phase and predicted topics through document clustering in the prediction phase \cite{bugert-etal-2021-generalizing}.

% \vspace*{-1mm}
\subsection{Training Phase}
% \vspace*{-1mm}
% \vspace*{-2mm}

\begin{figure}[b!]
  \centering
  \vspace*{-3mm}
\includegraphics[width=0.445\textwidth, center]{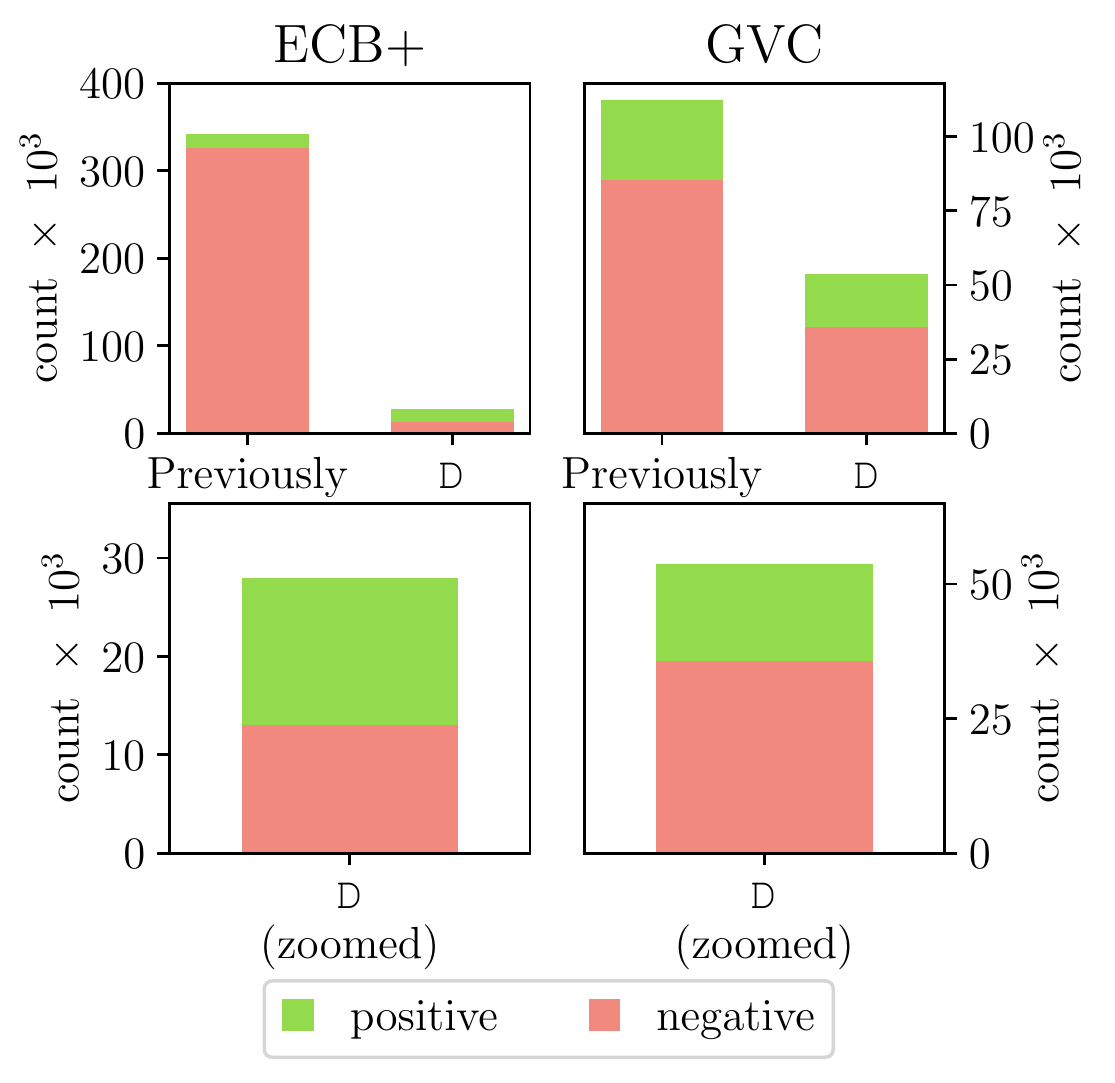}
\vspace*{-4mm}
\caption{Training Samples of previous methods vs. ours. The heuristic creates a balanced and significantly smaller training set for \ecb. For \gvc, the heuristic discards half of the negative samples while somewhat balancing the dataset.}  \label{fig:training-pairs}
\vspace*{-2mm}
\end{figure}

During the training phase, we leverage \LH~to generate a balanced set of positive and negative samples, labeled as \Ptp~and \Pfp, respectively. These samples are then used to train our models, \dSmall~and \dLong~separately, using the Binary Cross Entropy Loss (BCE) function as follows:
\begin{equation}
     L = \sum_{\substack{p_{+} \in~\Ptp, \\ p_{-} \in~\Pfp}}\log{\dPos(p_+)} + \log{(1 - \dPos(p_-))}\notag
 \end{equation}
Unlike traditional methods, we do not rely on random sampling or artificial balancing of the dataset. Instead, our heuristic ensures that the positive and negative samples are naturally balanced (as depicted in Figure \ref{fig:training-pairs}). A side-effect of adopting this approach is that some of the positive samples are excluded in training. We do this to keep the training and prediction phases consistent and, to ensure the cross-encoder is not confused by the inclusion of these hard positive examples.

Additionally, for \dPos~with Longformer, we utilize the entire document for training, while for \dPos~with RoBERTa, we only use the sentence containing the mention to provide contextual information. We employ the Adam optimizer with a learning rate of 0.0001 for the classifier and 0.00001 for fine-tuning the Transformer model. This entire process is illustrated in Algorithm \ref{alg:training}.
\begin{algorithm}[t]
\caption{Training Phase}\label{alg:training}
\begin{algorithmic}
\Require $D$: training document set
\Statex $T$: \text{gold topics}
\Statex $\Mv$: gold event mentions in $D$
\Statex $S^{v}$: sentences of the mentions
\Statex $D^{v}$: documents of the mentions
\Statex $G$: \text{gold mention cluster map}
\Statex {}
\State  $P \gets$ TopicMentionPairs$(\Mv, T)$
\State $\SynP \gets$ SynonymousLemmaPairs$(P, G)$

\State $\Ptp, \Pfp, \Pfn, \Ptn \gets \LH(P, G, \SynP, S^{v})$
\State $\dLong \gets$ TrainCrossEncoder(\Ptp, \Pfp, $D^{v}$)
\State $\dSmall \gets$ TrainCrossEncoder(\Ptp, \Pfp, $S^{v}$)
\State \Return $\SynP, \dLong, \dSmall$
\end{algorithmic}
\end{algorithm}

\newcommand{\Asubk}[1]{\ensuremath{\text{A}_{\text{#1}}}}
\newcommand{\AsubH}{\Asubk{H}}
\newcommand{\AsubP}{\Asubk{P}}
\newcommand{\AsubN}{\Asubk{N}}
\newcommand{\AsubPN}{\Asubk{PN}}

To ensure optimal performance, we train our system separately for both the \ecb~and \gvc~training sets. We utilize a single NVIDIA A100 GPU with 80GB memory to train \dLong~with the Longformer model, and a single NVIDIA RTX 3090 GPU (24 GB) for training \dSmall~with the RoBERTa-{\sc base} model.
\begin{algorithm}[t]
\caption{Prediction Phase}\label{alg:prediction}
\begin{algorithmic}
\Require $D$: testing document set
\Statex $T$: \text{gold/clustered topics}
\Statex $\Mv$: gold event mentions in $D$
\Statex $S^{v}$: sentences of the mentions
\Statex $\SynP$: synonymous lemma pairs from training
\Statex $\dSmall, \dLong$: trained \CE~discriminators
\Statex {}
\State  $P \gets$ TopicMentionPairs$(\Mv, T)$
\State $\AsubH, \Ppos \gets \LH(P, \SynP, S^{v})$
\State $\AsubP \gets \dSmall(\Ppos) > 0.5 $
\State $\AsubP \gets \dLong(\Ppos) > 0.5 $
\vskip1mm
\State \Return $\text{ConnectedComponents}(\AsubH)$,
\Statex \hskip3em  $\text{ConnectedComponents}(\AsubP)$
\end{algorithmic}
\end{algorithm}
We train each system for 10 epochs, with each epoch taking approximately one hour for the Longformer model and 15 minutes for the RoBERTa model.
\newcommand{\Stp}{$\mathcal{S}_{\text{TP}}$}
\newcommand{\Sfp}{$\mathsf{S}_{\text{FP}}$}
\newcommand{\Sfn}{$\mathcal{S}_{\text{FN}}$}
\newcommand{\Stn}{$\mathcal{S}_{\text{TN}}$}

% \Stp \Sfp

\vspace*{-2mm}

\subsection{Prediction Phase}
\vspace*{-1mm}
In the prediction phase, we first pass the mention pairs through the heuristic and create an adjacency matrix called \AsubH~based on its coreferent predictions. The ones predicted not coreferent by the heuristic are discarded. This step is crucial in terms of making the task tractable. Next, we pass the mention pairs that are predicted to be coreferent by the heuristic through \dSmall~and \dLong~separately. Using the subsequent coreferent predictions from these models, we generate another adjacency matrix \AsubP. To create event clusters, we use these matrices to identify connected components.

As a baseline, we use the matrix \AsubH~to generate the clusters. We then use \AsubP~to assess the improvements made by using \dSmall~and \dLong~over the baseline. This process is illustrated in  Algorithm \ref{alg:prediction}. The process takes between 6-10 minutes to run the Longformer model and between 1-2 minutes to run the RoBERTa one.

% \AsubH~\AsubP~\AsubN~\AsubPN

% \begin{figure}
%   \centering
%   \includegraphics[width=0.48\textwidth]{figures/prediction-phase-new.pdf}
%   \caption{In the prediction phase, the heuristic makes predictions about which mention pairs are coreferent and which are not. Mention pairs that are predicted to be coreferent are passed through \dPos, and those that are predicted to be not coreferent are discarded. \AsubH~and \AsubP~are adjacency matrices created using the coreferent predictions made by the heuristic and \dPos~respectively.}  \label{fig:PPhase}
% \end{figure}

%\subsection{Heuristic Threshold}

%\subsection{Clustering}

% \subsection{Negative Sampling for \dNeg}

% \newcommand{\dPosPlus}{${\texttt{D}^{+}}_{\texttt{POS}}$}
% \newcommand{\dNegPlus}{${\texttt{D}^{+}}_{\texttt{NEG}}$}
% \subsection{Continued pre-training (\dPosPlus, \dNegPlus)}

\section{Results}
\vspace*{-2mm}

\begin{table}[htb]
    \centering
    \resizebox{0.80\linewidth}{!}{
    \begin{tabular}{@{}lllccc@{}}
    \toprule
        &&& \multicolumn{3}{r}{CoNLL $F_1$} \\
        \cmidrule{4-6}
        &Methods&&  \ecb && \gvc \\
        \midrule
         & \citet{bugert-etal-2021-generalizing} && - && 59.4\\
         & \citet{cattan-etal-2021-cross-document} && 81.0 && - \\
         & \citet{caciularu-etal-2021-cdlm-cross} && 85.6 && -\\
         & \citet{held-etal-2021-focus} && {\bf 85.7} && {\bf 83.7} \\
        & \LH && 76.4 && 51.8 \\
         & \LH~+ \dSmall  && 80.3 && 73.7 \\
        & \LH~+ \dLong  && 81.7 && 75.0 \\
        \midrule
        & \LHOracle && 81.9 && 53.4 \\
        & \LHOracle~+ \dSmall && 85.9 && 75.4  \\
        & \LHOracle~+ \dLong && {\bf 87.4} &&\textbf{76.1}  \\
    \bottomrule
    \end{tabular}
    }
    % \vspace*{-2mm}
    \caption{Results on within and cross-document event coreference resolution on \ecb~and \gvc~test sets.}
    \label{tab:subtopic_results_event_gvc}
    \vspace*{-3.5mm}
\end{table}

We evaluate the event clusters formed using the standard coreference evaluation metrics (MUC, $B^{3}$, $CEAF_e$, LEA and CoNLL F1---the average of MUC, $B^{3}$ and $CEAF_e$ \citet{10.3115/1072399.1072405,Bagga98algorithmsfor, 10.3115/1220575.1220579, luo-etal-2014-extension, pradhan-EtAl:2014:P14-2,moosavi-etal-2019-using}). We run the baseline results (\LH~and \LHOracle) and the combination of each heuristic with the two discriminators (\LH/\LHOracle + \dSmall/\dLong). We compare to previous methods
 for \ecb~and \gvc~ as shown in Table \ref{tab:subtopic_results_event_gvc}. Bold indicates current or previous SOTA and our best model.

CoNLL F1 scores show that \LH~and \LHOracle~are strong baselines for the \ecb~corpus, where \LHOracle~surpasses some of the previous best methods. From this, we can say that making improvements in the heuristic by better methods of finding synonymous lemma pairs is a viable solution for tackling \ecb~with a heuristic. However, the heuristics fall short for \gvc, where \LHOracle~is only marginally better than \LH. This may be due to the lower variation in lemmas in the \gvc~corpus. We hypothesize methods that can automatically detect synonymous lemma pairs will not be beneficial for \gvc, and \LH~itself is sufficient as a heuristic here.

The discriminators consistently make significant improvements over the heuristics across both datasets. For \ecb, \dLong~is nearly 2 points better than \dSmall~in terms of the CoNLL measure. Both \dSmall~and \dLong~when coupled with \LHOracle~surpass the state of the art for this dataset. \LH~+\dLong~beats \citet{cattan-etal-2021-cross-document}~but falls short of SOTA, albeit by only 4 points. On \gvc, both fall short of SOTA \cite{held-etal-2021-focus} by only 8-9 points on CoNLL F1, with substantially fewer computations. In terms of computational cost-to-performance ratio, as we elaborate in \S \ref{sec:time}, our methods outperform all the previous methods.

%From the results, we also observe there is only a marginal difference between \dLong~and \dSmall. However, \dLong~is still slightly better than \dSmall. This indicates adding the whole context is not totally necessary. However, it might be interesting to find what exactly the model would consider as a beneficial context.

%\textcolor{red}{abhijnan's notes on negative results, feel free to exclude/trim}
For ECR, where context is key, we would expect better performance from encoders with longer context. \dLong~and \dSmall~show this trend for both \ecb~and \gvc~datasets. However, the gain we get from using the entire document is not substantial for the amount of additional computation required. An interesting line of future work would to automatically detect the core sections in the document that contribute to coreference and then only use that as context for ECR.

%\textcolor{red}{abhijnan's notes on negative results, feel free to exclude/trim}
% \vspace*{-2mm}
\section{Discussion}
% \vspace*{-2mm}
\subsection{Time Complexity Analysis}
\label{sec:time}
% \vspace*{-1mm}

\begin{figure}
  % \centering
  \scalebox{0.64}{\input{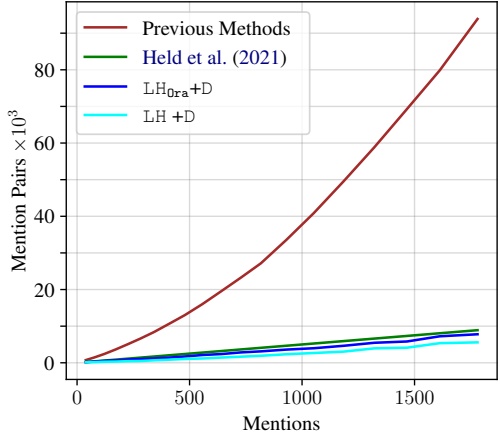}}
  \vspace*{-2mm}
\caption{Prediction Phase Time Complexity in terms of Mention Pair Encoding.}  \label{fig:test-pairs}
  \vspace*{-3.5mm}
\end{figure}
The heuristic is a very fast process that scales linearly with the number of mentions in a corpus. Specifically, by
hashing the lemma pairs and sentence token lemmas, this step performs linear comparisons of
mention pairs at prediction. The mention pair cross-encoding with Transformer is a computationally intensive process. A method that encodes all mention pairs in a large corpus can become intractable. Our method, however, is linear in complexity with the number of mentions, as shown in Figure \ref{fig:test-pairs}, and outperforms previous methods in terms of computational efficiency. While \citet{held-etal-2021-focus}'s cross-encoding at prediction is linear (5*n), their pruning step is quadratic. They rely additionally on training a bi-encoder and a mention neighborhood detector step that requires GPUs.
% \subsection{Synonymous Lemma Pairs Detector}

\vspace*{-1mm}
\subsection{Synonymous Lemma Pairs}
\vspace*{-1mm}
We have established an upper limit for ECR using the \LHOracle + \dLong~method for \ecb.  Previous methods such as \citet{held-etal-2021-focus}, use an oracle coreference scorer after their pruning step. In other words, their oracle assumption involves using a perfect cross-encoder. In contrast, we only use the oracle for pruning by assuming a perfect set of synonymous lemma pairs. This means that improved pruning methods can lead to better ECR performance.
% Our oracle assumption is weaker than using a perfect cross-encoder, and therefore, more extensible.
We believe that it is possible to create a more effective synonymous pair detector than \LHOracle~by adopting recent work on predicate class detection \cite{brown2014verbnet,brown2022semantic} that use VerbNet \cite{schuler2005verbnet}. In future research, we aim to enhance the process of generating synonymous pairs through the use of cross-encoding or additional steps such as word sense disambiguation with the Proposition Bank \cite{palmer-etal-2005-proposition, pradhan-etal-2022-propbank}. Identifying the sense of the trigger will help refine the lemma pairs that appear in coreference chains. Additionally, annotating the sense of the trigger is a straightforward process that can be easily incorporated into annotation procedures for new datasets, which is more efficient than coreference annotations.

\vspace*{-1mm}
\subsection{Qualitative Error Analysis}
\vspace*{-1mm}

We carry out a comprehensive analysis on errors the discriminator makes after the heuristic's predictions. Unlike previous methods \cite{barhom-etal-2019-revisiting} where they sample a subset of mentions to carry out the error analysis, we do so for the entire dataset. By efficiently discarding the large number of \Ptn,  we are able to isolate the shortcomings of the crossencoder, analyze them and offer solutions. Table \ref{tab:error_table_full} in Appendix \ref{sec:appendix_error} lists the various kinds of errors (incorrect and missing links) made by \dSmall~on the \ecb~and \gvc~dev sets.

We find error categories like same-sentence pronouns, weak temporal reasoning, ambiguity due to coreferring entities, misleading lexical similarity, and missed set-member coreferent links. Table~\ref{tab:error_table_full} in the appendix presents examples of each.

Incorrect links due to same-sentence pronouns like ``it'' and ``this'' can be avoided by refining the heuristics-based mention-pair generation process to exclude same-sentence pronouns. Similarly, ambiguous temporal contexts like "Saturday" and "New Year's Day" that refer to the day of occurrence of the same event in articles published on different dates can be resolved by leveraging more temporal context/metadata where available. Also, errors in lexically-different but semantically similar event mention lemmas can be reduced by leveraging more-enriched contextual representations.

By using the Oracle for pruning, we can focus on where \dSmall~falls short in terms of false positives. We first sort the final event clusters based on purity (number of non-coreferent links within the cluster compared to ground truth). Next, we identify pairs that the discriminator incorrectly predicted to be coreferent within these clusters, specifically focusing on highly impure clusters. We look for these pairs in highly impure clusters and analyze the mention sentences. Our findings are as follows:

\begin{itemize}
    \vspace*{-1mm}
    \item Problems caused when two big clusters are joined through very similar (almost adversarial) examples, e.g., ``British hiker'' vs. ``New Zealand hiker.'' This error can be fixed by performing an additional level of clustering, such as, K-means.
    \vspace*{-2mm}
    \item Problems with set-member relations, such as "shootings" being grouped with specific "shooting" events. The sets often include many non-coreferent member events. To address this issue, we can identify whether an event is plural or singular prior to coreference resolution. %This will allow us to more accurately group set events with their corresponding member events instead of corefering them with a coreference scorer.
    \vspace*{-2mm}
    \item Contrary to the notion that singleton mentions cause the most errors, we found that singletons appear in the {\it least} impure clusters. This means the cross-encoder discriminator is good in separating out singletons. %In contrast to the opinion that singletons cause the most errors, this is opposite of what we observed using a crossencoder.
\end{itemize}

% . Broadly, we go through the false negatives and the false positives by the \dSmall~on the development sets  of the two corpora.

%ToDo: list down representative examples. attach the entire xl file in appendix
\section{Conclusion \& Future work}
% \vspace*{-2mm}

We showed  that a simple heuristic paired with a crossencoder does  comparable ECR to more complicated methods while being computationally efficient. We set a upper bound for the performance on \ecb~suggesting improvement with better synonyms pairs detection we can achieve better results. Through extensive error analysis, we presented the shortcomings of the crossencoder in this task  and suggested ways to improve it.

Future research directions include applying our method to the more challenging task of cross-subtopic event coreference (e.g., FCC~\cite{bugert2020breaking}) where scalability and compute-efficiency are crucial metrics, making the current heuristic-based mention pair generation process ``learnable'' using an auxiliary cross-encoder, and incorporating word-sense disambiguation and lemma-pair annotations into the pipeline to resolve lexical ambiguity. An exciting direction for future work made tractable by our work is to incorporate additional cross-encoding features into the pipeline, especially using the latest advancements in visual transformers \cite{dosovitskiy2021image,bao2021BEiT,liu2021swin,radford2021learning}.  Another important direction is to test our method on languages with a richer morphology than English.

% Research directions:
% \begin{enumerate}
%     %\item Other datasets
%     %\item \DLH
%     %\item WSD into the pipeline using lexical resources
%     \item Lemma pair annotations
%     %\item Other languages
%     \item coreference core sections detection
% \end{enumerate}

%\section*{Acknowledgements}
% \newpage
\section*{Limitations}

The most evident limitation of this research is that is has only been demonstrated on English corefernce.  Using a lemma-based heuristic requires using a lemmatization algorithm in the preprocessing phase and for more morphologically complex languages, especially low-resourced ones, lemmatization technology is less well-developed and may not be a usable part of our pipeline. Application to more morphologically-rich languages is among our planned research directions.

In addition, all our experiments are performed on the gold standard mentions from \ecb~and \gvc, meaning that coreference resolution is effectively independent of mention detection, and therefore we have no evidence how our method would fare in a pipeline where the two are coupled.

A further limitation is that training of the cross-encoders still requires intensive usage of GPU hardware (the GPU used for training Longformer is particularly high-end).

%ACL 2023 requires all submissions to have a section titled ``Limitations'', for discussing the limitations of the paper as a complement to the discussion of strengths in the main text. This section should occur after the conclusion, but before the references. It will not count towards the page limit.
%The discussion of limitations is mandatory. Papers without a limitation section will be desk-rejected without review.

%While we are open to different types of limitations, just mentioning that a set of results have been shown for English only probably does not reflect what we expect.
%Mentioning that the method works mostly for languages with limited morphology, like English, is a much better alternative.
%In addition, limitations such as low scalability to long text, the requirement of large GPU resources, or other things that inspire crucial further investigation are welcome.

\section*{Ethics Statement}
%Scientific work published at ACL 2023 must comply with the ACL Ethics Policy.\footnote{\url{https://www.aclweb.org/portal/content/acl-code-ethics}} We encourage all authors to include an explicit ethics statement on the broader impact of the work, or other ethical considerations after the conclusion but before the references. The ethics statement will not count toward the page limit (8 pages for long, 4 pages for short papers).

We use publicly-available datasets, meaning any bias or offensive content in those datasets risks being reflected in our results.  By its nature, the Gun Violence Corpus contains violent content that may be troubling for some.

We make extensive use of GPUs for training the discriminator models as part of our pipeline.  While this has implications for resource consumption and access implications for those without similar hardware, the linear time complexity of our solution presents a way forward that relies less overall on GPU hardware than previous approaches, increasing the ability to perform event coreference resolution in low-compute settings.

\section*{Acknowledgements}
We would like to express our sincere gratitude to the anonymous reviewers whose insightful comments and constructive feedback helped to greatly improve the quality of this paper. We gratefully acknowledge the support of U.S. Defense Advanced Research Projects Agency (DARPA) FA8750-18-2-0016-AIDA – RAMFIS: Representations of vectors and Abstract Meanings for Information Synthesis. Any opinions, findings, and conclusions or recommendations expressed in this material are those of the authors and do not necessarily reflect the views of DARPA or the U.S. government. Finally, we extend our thanks to the BoulderNLP group and the SIGNAL Lab at Colorado State for their valuable input and collaboration throughout the development of this work.

% This research was supported in part by a contract on grant award FA8750-18-2-0016 from the U.S. Defense Advanced Research Projects Agency (DARPA). Views expressed herein do not reflect the policy or position of the Department of Defense or the U.S. Government. All errors are the responsibility of the authors.
\bibliography{bibs/easy_first, bibs/datasets, bibs/anthology, bibs/coreference,bibs/image}
\bibliographystyle{acl_natbib}

\appendix
\section{Ablation Study of Global Attention}
\label{sec:appendix_ablation}

\begin{table}[!htb]
\centering
 \begin{tabular}{||c c c||} 
 \hline
{\bf Features} & {\bf ECB+} & {\bf GVC}  \\ [0.5ex] 
 \hline\hline
        \bf {w/o global attn.}  & 85.0 & 76.5\\ 
        \bf {w/ global attn.} & 82.9 & 77.0 \\

 \hline
 
 \end{tabular}

  \vspace*{-2mm}
 \caption{Table showing the CoNLL F1 scores from the \dPos~Encoder with and without Longformer Global Attention on \gvc~and \ecb~dev sets. }
\label{table:ablation_global}
\vspace*{-2mm}
\end{table}

% \begin{table}[!htb]
% \centering
%     \begin{tabular}{llll}
% \toprule
% \small{\bf Features} & \small{\bf ECB+} & \small{\bf GVC   \\
%         \cmidrule(lr){1-3}
%         \small{{\bf W/O Global}}  & 85.0 & 76.5\\ 
%         \small{{\bf W-Global}} & 82.9 & 77.0 \\
%  \end{tabular}
%  \vspace*{-2mm}
%  \caption{Table showing the CoNLL F1 scores from the Dpos Encoder with and without Longformer Global Attention on the GVC and the ECB dev set. }
% \label{table:ablation_global}
% \vspace*{-2mm}
% \end{table}

Table~\ref{table:ablation_global} compares \dLong~performance with and without Longformer global attention on the \ecb~and \gvc~dev sets. This shows a dataset-specific contrast vis-{\`a}-vis sequence length where performance with global attention on \gvc~dev set is only \textit{marginally} better than without, while the reverse is seen on the \ecb~dev set. More specifically, this suggests that perhaps the ``relevant'' or "core" context for ECR lies closer to the neighborhood of event lemmas (wrapped by trigger tokens) than the CLS tokens (that use global attention) in both corpora, albeit more so in \ecb. As such, applying global attention to the CLS tokens here encodes more irrelevant context. Therefore, \dLong~with Longformer global attention performs less well on \ecb~while being almost comparable to \dLong~without global attention on \gvc.
\section{Full Results}
\label{sec:appendix_full}

Table~\ref{tab:subtopic_results_event_gvc_full} shows complete results for all metrics from all models for within and cross-document coreference resolution on the \gvc~test set.  Table~\ref{tab:subtopic_results_event_full} shows complete results for all metrics from all models on the \ecb~test set.

\begin{table*}[!ht]
    \centering
    \resizebox{\textwidth}{!}{
    \begin{tabular}{@{}lllrrrrrrrrrrrrrrrrr@{}}
    \toprule
    &&& \multicolumn{3}{c}{MUC} && \multicolumn{3}{@{}c@{}}{$B^3$} & & \multicolumn{3}{c}{$CEAFe$} && \multicolumn{3}{c}{LEA} && CoNLL\\
    \cmidrule{4-6} \cmidrule{8-10} \cmidrule{12-14} \cmidrule{16-18} \cmidrule{20-20}
    &&& R & P & $F_1$ && R & P & $F_1$ && R &P & $F_1$ && R &P & $F_1$ && \multicolumn{1}{r}{$F_1$}  \\
   \midrule

        & \citet{bugert-etal-2021-generalizing} && 78.1 & 66.3 & 71.7&& 73.6 &49.9& 59.5 &&38.2& 60.9& 47.0&&  56.5 &38.2& 45.6 &&59.4\\
        & \citet{held-etal-2021-focus} && 91.8 &91.2 & \textbf{91.5} && 82.2& 83.8& \textbf{83.0} && 75.5 & 77.9& \textbf{76.7} && 79.0 & 82.3 & \textbf{80.6} && \textbf{83.7} \\
        & \LH && 94.8  & 82.0 & 87.9 && 90.1 & 28.5 & 43.3 && 16.3 & 47.8 & 24.3 && 85.1 & 23.9 & 37.4 && 51.8  \\
        & \LHOracle && 95.2  & 82.3 & 88.3 && 91.2 & 29.1 & 44.1 && 18.6 & 54.7 & 27.8 && 86.4 & 24.9 & 38.6 && 53.4 \\
         & \LH~+ \dSmall  && 87.0  & 89.6 & 88.3 && 82.3 & 67.9 & 74.4 && 62.0 & 55.2 & 58.4 && 77.6 & 57.8 & 66.2 && 73.7 \\
         % & \DLH~+ \dPos (roberta-base) && -  & - & - && -& - & - && - & - & - && - & - & - && - \\

        & \LHOracle~+ \dSmall && 89.1  & 90.2 & \textbf{89.6} && 85.0 & 68.0 & 75.6 && 62.7 & 59.6 & 61.1 && 80.6 & 59.5 & 68.5 && 75.4  \\
        & \LH~+ \dLong && 84.0  & 91.1 & 87.4 && 79.0 & 76.4 & 77.7 && 69.6 & 52.5 & 59.9 && 74.1 & 63.9 & 68.6 && 75.0 \\
        & \LHOracle~+ \dLong && 84.9  & 91.4 & 88.0 && 80.4 & 77.4 & 78.9 && 70.5 & 54.3 & \textbf{61.3} && 75.7 & 65.5 & \textbf{70.2} && \textbf{76.1} \\
    \bottomrule
    \end{tabular}}
    \caption{Results on within and cross-document event coreference resolution on \gvc~test set. Bolded F1 values indicate current or previous state of the art according to that metric as well as our best model.}
    \label{tab:subtopic_results_event_gvc_full}
    % \vspace{-3mm}
\end{table*}
\begin{table*}[!ht]
    \centering
    \resizebox{\textwidth}{!}{
    \begin{tabular}{@{}lllrrrrrrrrrrrrrrrrr@{}}
    \toprule
    &&& \multicolumn{3}{c}{MUC} && \multicolumn{3}{@{}c@{}}{$B^3$} & & \multicolumn{3}{c}{$CEAFe$} && \multicolumn{3}{c}{LEA} && CoNLL\\
    \cmidrule{4-6} \cmidrule{8-10} \cmidrule{12-14} \cmidrule{16-18} \cmidrule{20-20}
    &&& R & P & $F_1$ && R & P & $F_1$ && R &P & $F_1$ && R &P & $F_1$ && \multicolumn{1}{r}{$F_1$}  \\
   \midrule

        & \citet{barhom-etal-2019-revisiting} && 78.1 & 84.0 & 80.9 && 76.8 & 86.1 & 81.2 && 79.6 & 73.3 & 76.3 && 64.6 & 72.3 & 68.3 && 79.5\\

        & \citet{meged-etal-2020-paraphrasing} && 78.8 & 84.7 & 81.6 && 75.9 & 85.9 & 80.6 && 81.1 & 74.8 & 77.8 && 64.7 & 73.4 & 68.8 && 80.0\\

       & \citet{cattan-etal-2021-cross-document} &&  85.1 & 81.9 & 83.5 && 82.1 & 82.7 & 82.4 && 75.2 & 78.9 & 77.0 && 68.8 & 72.0 & 70.4 && 81.0 \\

     &\citet{zeng-etal-2020-event} &&  85.6 & 89.3 & 87.5 && 77.6 & 89.7 & 83.2 && 84.5 & 80.1 & 82.3  && - & - & - &&  84.3   \\

      &   \citet{yu2022pairwise} &&  88.1 & 85.1 & 86.6 && 86.1 & 84.7 & 85.4 && 79.6 & 83.1 & 81.3 && - & - & - &&  84.4   \\

         &\citet{allaway-etal-2021-sequential} &&  81.7 & 82.8 & 82.2 && 80.8 & 81.5 & 81.1 && 79.8  & 78.4 & 79.1  && - & - & - &&  80.8 \\

        & \citet{caciularu-etal-2021-cdlm-cross} && 87.1  & 89.2 & \textbf{88.1} && 84.9 & 87.9 & 86.4 && 83.3 & 81.2 & 82.2 && 76.7 & 77.2 & \textbf{76.9} && 85.6 \\

        % 87.0 & 88.1 87.5 85.6 87.7 86.6 80.3 85.8 82.9 85.7 74.9 73.2 74.0
        & \citet{held-etal-2021-focus} && 87.0  & 88.1 & 87.5 && 85.6 & 87.7 & \textbf{86.6} && 80.3 & 85.8 & \textbf{82.9} && 74.9 & 73.2& 74.0&& \textbf{85.7} \\

        & \LH && 85.1  & 75.6 & 80.1 && 83.2 & 72.2 & 77.3 && 66.2 & 78.1 & 71.7 && 67.3 & 62.6 & 64.9 && 76.4  \\

        & \LHOracle && 99.1 & 79.6 & 88.3 && 97.9 & 67.7 & 80.0 && 65.9 & 93.7 & 77.4 && 85.1 & 63.8 & 72.9 && 81.9 \\

        & \LH~+ \dSmall && 76.2  & 86.9 & 81.2 && 77.8 & 85.7 & 81.6 && 83.9 & 73.0 & 78.1 && 68.7 & 71.5 & 70.1 && 80.3 \\

        & \LHOracle~+ \dSmall && 89.8 & 87.6 & 88.7 && 90.7 & 80.2 & 85.1 && 82.5 & 85.1 & 83.8 && 83.3 & 72.2 & 77.3 && 85.9 \\
         % & \DLH~+ \dPos && -  & - & - && -& - & - && - & - & - && - & - & - && - \\

        & \LH~+ \dLong && 80.0  & 87.3 & 83.5 && 79.6 & 85.4 & 82.4 && 83.1 & 75.5 & 79.1 && 70.5 & 73.3 & 71.9 && 81.7  \\

        & \LHOracle~+ \dLong && 93.7 & 87.9 & \textbf{90.7} && 94.1& 79.6 & \textbf{86.3} && 81.6 & 88.7 & \textbf{85.0} && 86.8 & 73.2 & \textbf{79.4} && \textbf{87.4} \\

    \bottomrule
    \end{tabular}}
    \caption{Results on  within and cross-document event coreference resolution on ECB+ test set with gold mentions and predicted topics. Bolded F1 values indicate current or previous state of the art according to that metric as well as our best model.}
    \label{tab:subtopic_results_event_full}
    % \vspace{-3mm}
\end{table*}

\section{Qualitative Error Examples}
\label{sec:appendix_error}

Table~\ref{tab:error_table_full} presents an example of each type of error we identified in the output of our discriminator (\dSmall).

\begin{table*}[!ht]
\centering%
%\caption{}
\begin{tabularx}{\linewidth}{lX}
\toprule
{\small Category} & {\small Snippet} \\
\midrule
\midrule

{\small Adversarial/Conflicting } & {\small British climber <m> dies </m> in
New Zealand fall.....The first of the <m>\underline{ deaths} </m> this weekend was that of a New Zealand climber who fell on Friday morning. }\\

{\small Adversarial/Conflicting } & {\small British climber <m> dies </m> in New Zealand fall.....Australian Ski Mountaineer <m>\underline{ Dies}</m> in Fall in New Zealand. }\\

{\small Adversarial/Conflicting } & {\small ..Prosecutor Kym Worthy announces charges against individuals involved in the gun violence <m> deaths </m> of children in Detroit ..... Grandparents charged in 5-year - old 's shooting <m> \underline{death} </m> Buy Photo Wayne County Prosecutor Kym Worthy announces charges against individuals involved in the gun violence deaths of children... }\\

{\small Pronoun Lemmas} & {\small This just does not happen in this area whatsoever . <m> It </m>’s just unreal , ” said neighbor Sheila Rawlins....<m> \underline {This} </m> just does not happen in this area whatsoever . It ’s just unreal , ” said neighbor Sheila Rawlins . }\\

{\small Set-Member Relationship} & {\small On Friday , Chicago surpassed 700 <m> homicides </m> so far this year .
 ....<m> \textbf{Homicide} </m> Watch Chicago Javon Wilson , the teenage grandson of U.S. Rep. Danny Davis , was shot to death over what police called an arugment over sneakers in his Englewood home Friday evening .}\\

{\small Weak Temporal Reasoning} & {\small Police :  in an unrelated <m> incident </m> a man was shot at 3:18 a.m. \underline{Saturday} in North Toledo ....Toledo mother grieves 3-year - old 's <m> \textbf{shooting}</m> death | Judge sets bond at 580,000 USD for Toledo man accused of rape , kidnapping | Toledo man sentenced to 11 years in \underline{New Year 's Day} shooting}\\

{\small Incomplete, Short Context} & {\small Ellen DeGeneres to <m> Host </m> Oscars....It will be her second <m> \textbf {stint} </m> in the job , after hosting the 2007 ceremony and earning an Emmy nomination for it .}\\

{\small Similar context, Different event times} & {\small  near Farmington Road around \underline {9 p.m.} There they found a 32-year - old unidentified man with a <m> gunshot </m> wound outside of a home ....The family was driving about \underline{8:26 p.m.} Sunday in the 1100 block of South Commerce Street when <m> \underline{gunshots were fired }</m> from a dark sedan that began following their vehicle... }\\

{\small Same Lemma, Ambiguous Context} & {\small Police : Man Shot To Death In Stockton Related To 3-Year - Old <m> Killed </m> By Stray Bullet 2 p.m. UPDATE : Stockton Police have identified the man shot and killed on ....Police : Man Shot To Death In Stockton Related To 3-Year - Old Killed By Stray Bullet 2 p.m. UPDATE : Stockton Police have identified the man shot and <m> \underline{killed} </m> on Tuesday night. }\\

{\small Lexically different, Semantically same} & {\small One man is dead after being <m> shot </m> by a gunman ....Employees at a Vancouver wholesaler were coping Saturday with the death of their boss , who was <m> \textbf{gunned down} </m> at their office Christmas party .}\\

{\small Misc.} & {\small Baton Rouge Police have charged 17-year - old Ahmad Antoine of Baton Rouge with Negligent Homicide in the city ’s latest shooting <m> death </m> .....Tagged Baton Rouge , <m> \underline{homicide} </m>.}\\

% \midrule
% A & B \\
\bottomrule
\end{tabularx}

\caption {Qualitative Analysis on the hard mention pairs incorrectly linked (or missed) by our Discriminator (\dSmall) in the ECB+ and GVC dev set: Underlined and bold-faced mentions surrounded by trigger tokens respectively indicate incorrect and
missing assignments. Underlined spans without trigger tokens represents the category-specific quality being highlighted. The miscellaneous category (Misc.) refers to other errors including
(reasonable) predictions that are either incorrect annotations in the gold data or incomplete gold sentences. }
    \label{tab:error_table_full}
\end{table*}

% \begin{table*}[!ht]
%     \centering
%     \begin{center}
% \begin{tabular}{ |c|cccccccc| }
%  \hline
%   {\small Pronouns \linebreak
%   resolution} & {\small This just does not happen in this area whatsoever . <m> It </m> ’s just unreal , ” said neighbor Sheila Rawlins....\linebreak.<m> This </m> just does not happen in this area whatsoever . It ’s just unreal , ” said neighbor Sheila Rawlins .}
%   \\
%   \hline
%  \\
%  \hline
% \end{tabular}
% \end{center}
%     \caption{Results on within and cross-document event coreference resolution on \gvc~test set.}
%     \label{tab:subtopic_results_event_gvc}
% \end{table*}

\end{document}